# COMPARISON AND ANALYSIS OF PHOTO IMAGE FORGERY DETECTION TECHNIQUES

S.Murali[1], Govindraj B. Chittapur[2], Prabhakara H. S[1] and Basavaraj S. Anami[3]

[1]Mahraja Institute Of Technology, Mysore, INDIA
nymurali@yahoo.com
prabhassan@gmail.com
[2]Basaveshwar Engineering College, Bagalkot, INDIA
govindraj_bec@yahoo.com
[3] KLE Institute Of Technology, Hubli, INDIA
anami_basu@hotmail.com

## ABSTRACT

*Digital Photo images are everywhere, on the covers of magazines, in newspapers, in courtrooms, and all over the Internet. We are exposed to them throughout the day and most of the time. Ease with which images can be manipulated; we need to be aware that seeing does not always imply believing. We propose methodologies to identify such unbelievable photo images and succeeded to identify forged region by given only the forged image. Formats are additive tag for every file system and contents are relatively expressed with extension based on most popular digital camera uses JPEG and Other image formats like png, bmp etc. We have designed algorithm running behind with the concept of abnormal anomalies and identify the forgery regions.*

## 1. INTRODUCTION

The art of making an image forgery is almost as old as photography itself. In its early years, photography quickly became the chosen method for making portraits, and portrait photographers learned that they could improve sales by retouching their photographs to please the sitter [1]. Photo manipulation has become more common in the age of digital cameras and image editing software. Gathered below are examples of some of the notable instances of photo manipulation in history. so we focus here on the instances that have been most controversial or notorious, or ones that raise the most interesting ethical questions [2]. The photographers have also experimented with composition, i.e., combining multiple images into one. An early example of composition appears in the figures 1.1 and 1.2.

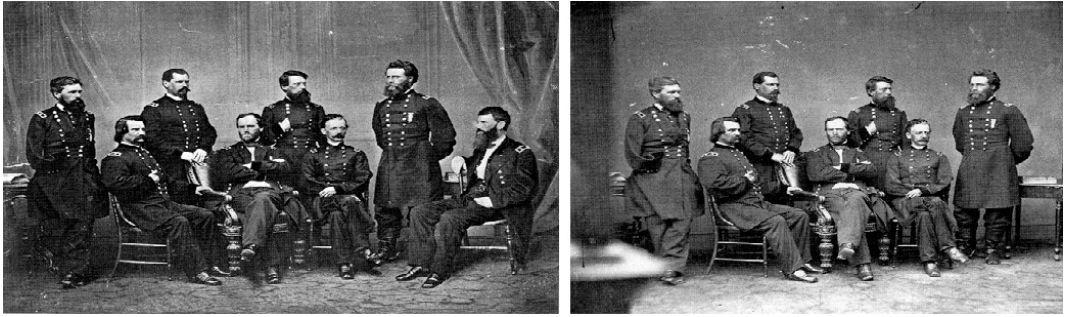

Fig. 1.1: 1865: In this photo by famed photographer Mathew Brady, General Sherman is seen posing with his Generals. General Francis P. Blair (far right) was added to the original photograph. The photo on the left is another image from the same sitting, at which General Blair was not in attendance.

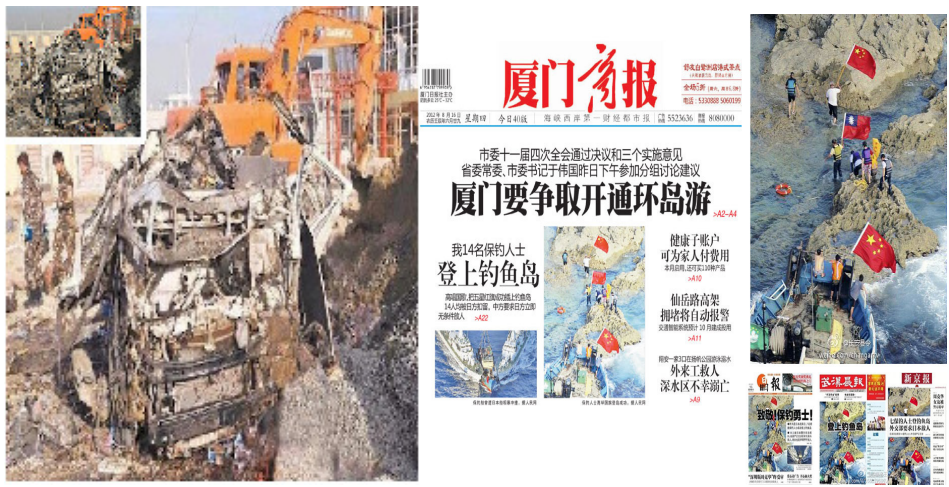

Fig 1.2 Recent Forgery Photos published in popular news media and magazines

Digital images offer many attributes for tamper detection algorithm to take advantage of - specifically the color and brightness of individual pixels as well as an image's resolution and format. These properties allow for analysis and comparison between the fundamentals of digital forgeries in an effort to develop an algorithm for detecting image tampering. This paper focuses on images saved in the JPEG format. Therefore a research work on basis of compression scheme is discussed to determine what information can be gathered about a digital forgery saved in this format. Other fundamental properties of any digital forgery are used to develop additional detection technique such as direction filter, which is used to detect the forgery region when we conduct the experiments on gray level of photos.

In this paper, we propose a novel scheme for identifying the location of copy-create and copy-move supported tampering algorithms and authenticating an Image by applying the JPEG Block and Direction Filter Techniques.

This paper is organized as follows. After reviewing related works in Section II, we present basic ideas of related photo forgery techniques, and details of the proposed method of identifying and computing the prediction of precession rate and recall rate tampered regions in an image where images are altered with image editing software like paint and Photoshop editor along with different image format supported by modern digital cameras in Section III. The experimental results and system boundary discussion were present in Section IV. Final conclusions are drawn in Section V.

## 2. RELETED WORK ON PHOTO IMAGE FORGERY

This section introduces the techniques and methods currently available in the area of digital image forgery detection. Currently, most acquisition and manipulation tools use the JPEG standard for image compression. As a result, one of the standard approaches is to use the blocking fingerprints introduced by JPEG compression, as reliable indicators of possible image tampering. Not only do these inconsistencies help to determine possible forgery, but they can also be used to light into what method of forgery was used. Many passive schemes have been developed based on these fingerprints to detect Resampling [4], Copy-paste [5,6], Luminance-level [7,], Double Compression JPEG [8,16], ANN [9], and Wavelet Transformation Coefficient [10].Above mentioned methodologies are derived from one another and they all contain constraints in implementations and limitations in performance. We concentrate on media photo images and propose and to develop an effective algorithm for detecting the forgery region in most popular image format JPEG and other digital camera supported image formats.

Detection of digital image forgery having enormous number applications related Forensic science document questioning section although which is very helpful for media, publication, law, military, Medical image science application, satellite image, research and World Wide Web publications.

## 3. METHODOLOGY AND IMPLEMENTATION OF DETECTING DIGITAL IMAGE FORGERY

Photo image forgery is classified in to two categories. The first class of image forgeries includes images tampered by copying one area in an image and pasting it onto another area. It is called as Copy-Move Forgery or Cloning. The second class of forgeries is copying and pasting areas from one or more images and pasting on to an image being forged. The image processing community formally refers to this type of image as an image "composition," which is defined as the "digitally manipulated combination of at least two source images to produce an integrated result". It is also called as Copy-Create Image Forgery.

We have developed effective methodologies for detecting both Copy-Move and Copy-Create type of image forgeries.

### 3.1 Methodology Based On JPEG Compression Analysis and Algorithm for Forgery Detection

Block-based processing is a popular technique used in image processing where the image is broken into sub-parts or equal squares. Each block is considered as a sub-image. This method is allows recursive type processing, with the sub-processing resembling a "divide and conquer" approach. Block-based processing is useful because the calculations performed are influenced by only the information present in that particular block.Block-based processing is employed in image compression. JPEG compression is block-DCT based, and a popularly used image compression technology. The compression standard set forth by the International Standards Organization (ISO) and International Electro-Technical Commission (IEC) of Joint Photographic Expert Group (JPEG) images uses a Discrete Cosine Transform (DCT) scheme [11]. A simplified diagram of the JPEG compression is illustrated in Fig. 3.1

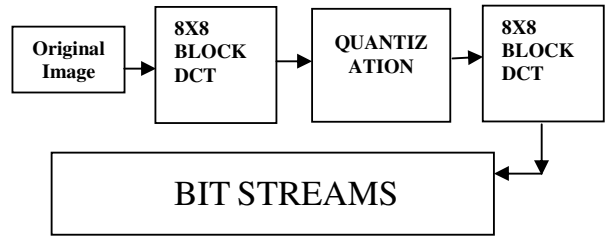

Fig 3.1 Block Diagram JPEG Compression

The DCT domain is used to convert a signal into coefficient values with the ability to perform truncating and rounding operations, thus allowing compression of this signal to take place. The JPEG compression process starts by calculating the DCT of each 8x8 blocks in the image based on the following formula [12]:

$$D_{i,j} = \sum_{k,l=0}^{7} a_{kl}(i,j) B_{kl}, \text{------------------}(1)$$

Where where $\frac{1}{4} w(k) w(l) \cos \frac{\lambda}{16} k(2j+1)$ and $w(k) = \frac{1}{\sqrt{2}}$ for k=0 and w(k)=1 otherwise Matrix **D**, which contains 64 DCT coefficients, is then quantized using a quantization matrix Q [12]:

$$D_{i,j} = \text{round} \left( \frac{D_{i,j}}{Q_{i,j}} \right) i,j \in \{0,1,2,3,4,5,6,7\} \text{----------}(2)$$

The quantized coefficients, Dij, are then arranged in a zigzag order, encoded using the Huffman Algorithm, and inserted into what makes up the JPEG file [13]. Decomposition works similarly just in reverse order. By rounding the ratio above, an integer value is obtained and thus allows an image to be compressed. A threshold is set to determine what integer values should effectively be discarded. The parts to be discarded are carefully calculated based on a "Quality Factor", which is a reference number between 0 and 100 [13]. The higher the Quality Factor, the less compressed and the better quality the image is. A trade-off between file size and image quality is always necessary in this type of lossy compression.

A JPEG image can either be color or grayscale. The above operations encode pixel values that are usually in the 0 to 255 range (8-bits). In the case of grayscale images, a sole 8-bit number represents the level of gray in each pixel. Color images use similar boundaries but include three 8-bit numbers, one for each of the Red, Green, and Blue channels. This allows for the creation of a 24-bit color image [14]. The analysis in this section works for all types of JPEG images and the various forensics approaches apply regardless of the color type.

Whenever an image is compressed using the JPEG scheme, a distinct phenomenon occurs. The 8x8 blocks, resulting from the DCT function and subsequent information loss, become easily noticeable. The blocks are easily distinguishable in this image and show the effects of DCT compression.With a somewhat predictable scheme used by the JPEG compression algorithm, the analysis of an image with respect to this scheme may show promise in detecting image tampering. JPEG compression forms a type of "fingerprint" that may indicate alteration.

---

Step 1: Divide image into disjoint 8x8 compression blocks (i,j) for each 8x8 JPEG compression block (i,j) within bounds:

$$R(i, j) = |A - B - C - D| \quad \text{---------------(1)}$$

where $A = pixelValue(8*i, 8*j)$, $B = pixelvalue(8*i, [8*j]+1)$, $C = pixelvalue([8*i]+1, 8*j)$, $D = Pixelvalue([8*i]+1, [8*j]+1)$

Step 2: For each 8x8 JPEG compression $Block(i,j)$ within bounds

$Dright(i, j) = |R(i, j) - (R(i, j+1))|$

$Dbottom(i, j) = |R(i, j) - (R(i+1, j))|$

Step 3: For each *8x8* JPEG compression block *(i,j)* within bounds

    **If** $(Dright(i, j) \geq t)$ or $Dbottom(i, j) \leq t)$

       set all pixel values in *(i,j)* to white

**else**

       Set all pixel values in *(i,j)* to black

    **End**

---

Fig 3.2 JPEG Image Compression Algorithm is given below:

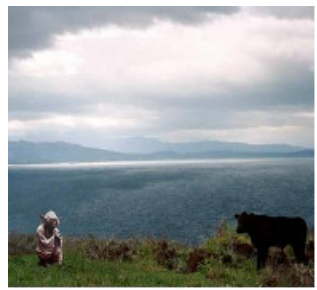
(a) Forged Image

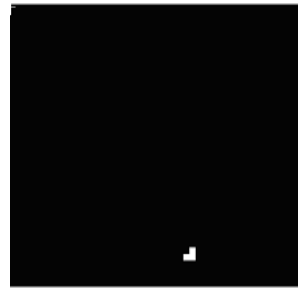
(b) Threshold = 75

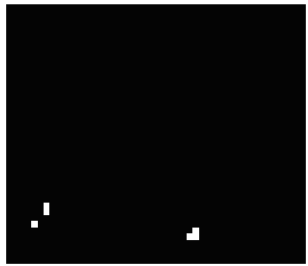
(c ) Threshold =65

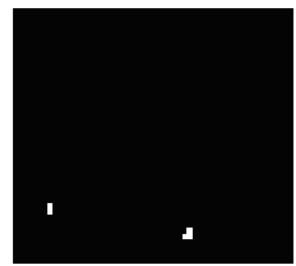
(d) Threshold =55

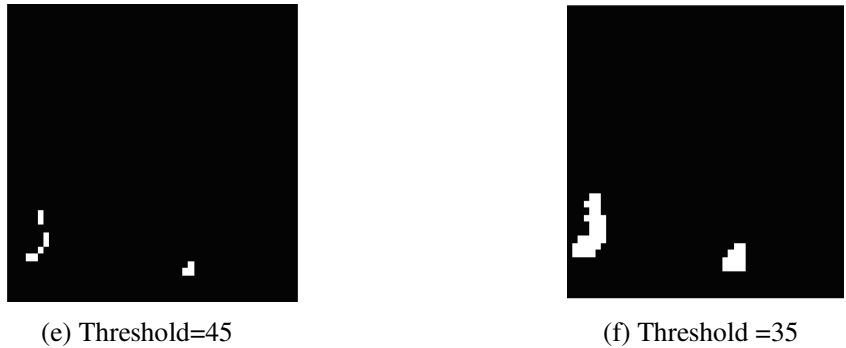

(e) Threshold=45          (f) Threshold =35

Fig 3.3 Random Selection of Global Threshold Setting and Evaluation used in Jpeg compression algorithm for Photo Forgery Detection.

## 3.2 Method Based on Direction Filter Using JPEG Image Analysis

General forgery detection methods are based on JPEG compression threshold which work for only JPEG image format. Today digital cameras support other image formats also. For this reason we propose novel methodology for photo forgery detection based on standard deviation based edge detection that detects the edges present in all directions. The main steps of proposed algorithms are based on Image Edge Detection and tampering localization. Following Steps explain the process of forgery detection:

**Step 1: Image Pre-processing:** If the image data is not represented in HSV color space, it is converted to this color space by means of appropriate transformations. Our system only uses the intensity data (v-channel of HSV) during further processing. Here V Channel represents the intensity image.

$$I(\lambda) = \rho(\lambda) L(\lambda) \quad \text{------------} \quad (1)$$

where $L(\lambda)$ spectral Enery distrubtion with $\lambda$ wavelength $p(\lambda)$ reflactivity of the object. The Luminance of spacially distrubuted object with light distrubution object with light distrubtion $I(x, y, \lambda)$ is defined as:

$$f(x, y) = \int_0^\infty I(x, y, \lambda) V(\lambda) d\lambda \quad \text{----------} \quad (2)$$

where $v(\lambda)$ is relative luminance efficiency function of the system.

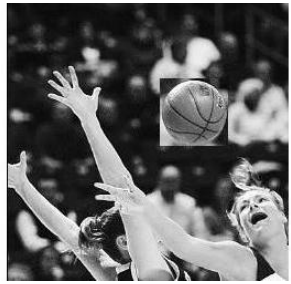 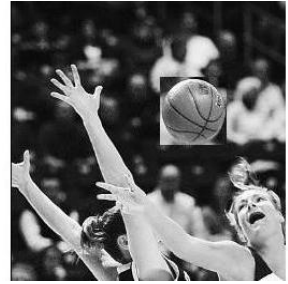

(a) Forged Image    (b) V-Channel

**Step 2**: **Edge Detection:** This step focuses the attention to areas where tampering may occur. We employ a simple method for converting the gray-level image into an edge image. Our algorithm is based on the fact that the tampered image region possesses high standard deviation surrounds the tampered region.

$$Std(x) = \frac{1}{(N-1)} \sum (w(I) - \mu(x)) \;\text{---------------} (3)$$

If W(I) where x is a set of all pixels in a sub-window µ(x), N is a number of pixels in W(i), µ(x) is mean value of V(I)and I€ W(i). A window size of 3x7 pixels was used in this step:

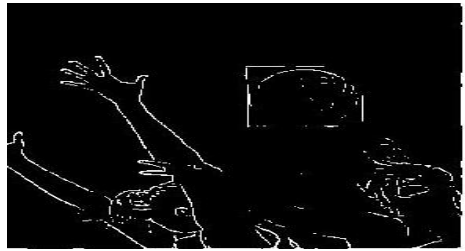

(c) Edge Detection

**Step3: Localization of tampered part:** Horizontal and Vertical projections are calculated and with the help of horizontal and vertical thresholds other directional edges are removed. Horizontal and Vertical edges images are combined together and feature map is generated.

$$H - Threshold = \text{Mean(Horizontal Projection)} \;\text{-----------------} (4)$$
$$V - Threshold = \text{Mean(Vertical Projection)} \;\text{--------------------} (5)$$

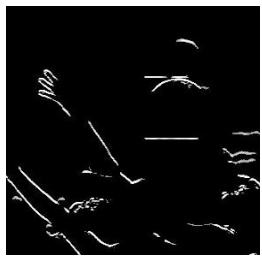 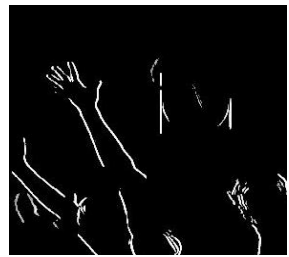

(d) Horizontal Projection    (e) Vertical Projection

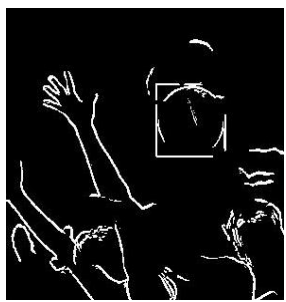 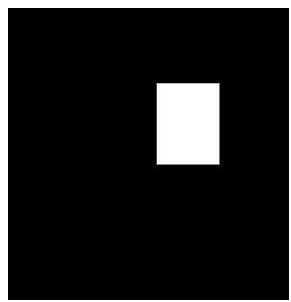

(f) Combined Projection          (g) Feature Map

Feature Map: It is a binary image same size as original image where high intensity indicates possibility of tampering.

## 4 EXPERIMENTAL RESULTS AND DISCUSSION

The proposed approach has been evaluated using datasets containing different types of tampered images. The test data consists of 100 images.

**Case I: Results of JPEG Block Technique:** Two images depicting a helicopter in the sky are taken and

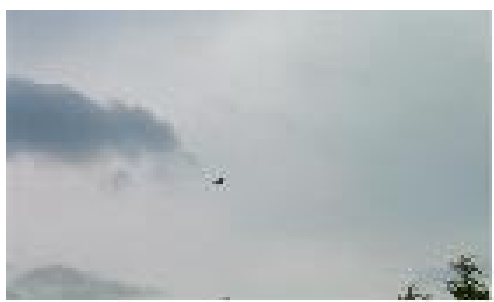 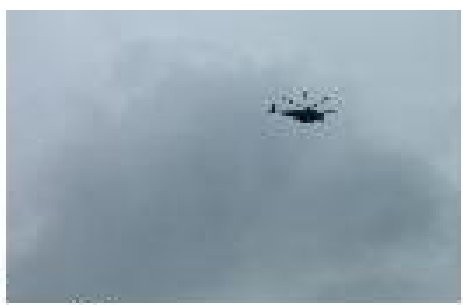

(a)                              (b)

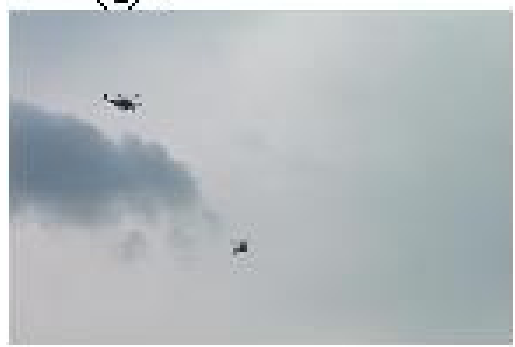 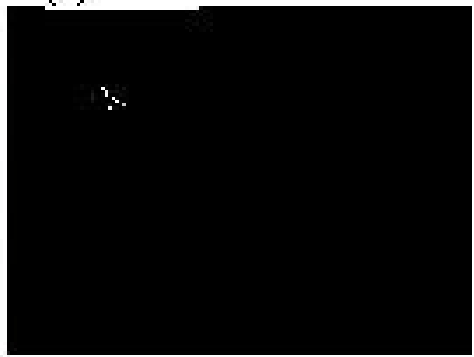

(c) Forged Image From (a) (b)     (d) Tampared Object

International Journal on Computational Sciences & Applications (IJCSA) Vo2, No.6, December 2012

Figure 3.5: From figures (a) and (b) forged image (c) is created using Photoshop. Results of tampered object detection are shown in (d).

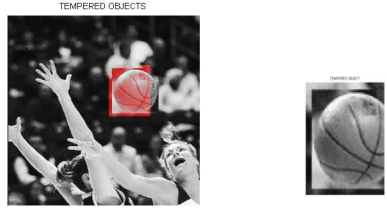

```
1: Read Input Image
2: Extract edge image using CANNY EDGE Operator
3: Computer X and Y-axis pixel by applying convolution with directional filter to
      GENERATE SIMILAR type of pixel pyramid.
4: Calculate Horizontal and Vertical projection profile
5: Find boundary pixel values, which differ with Projection profile with X, and Y values
6:   Calculate feature map
7: Identify the forgery region
8: Display the Forgery Region
9: Extract the forgery Region
```

Fig 3.4: Algorithm for forgery detection using direction filter.

**Case II: Result of Direction Filter:** The proposed approach has been evaluated using datasets containing different types of tampered images. The test data consists of 100 images

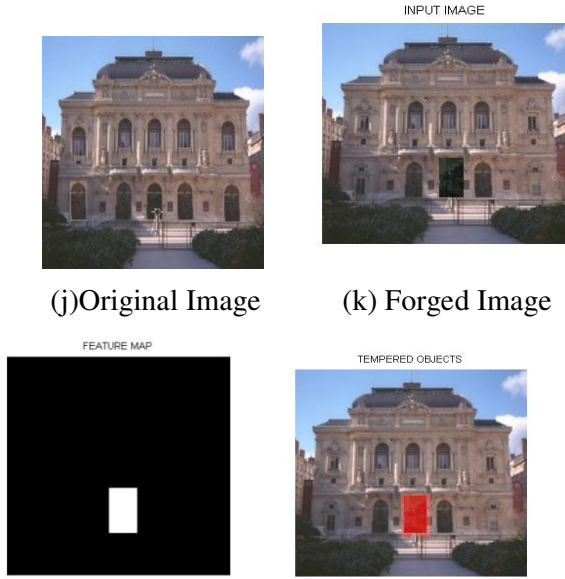

(j)Original Image      (k) Forged Image

(l) Feature Map      (m) Marked Tampered  From the Equations (6) and (7), The precision and recall rates have been computed based on the number of correctly detected tampered parts in an image in order to evaluate the efficiency and robustness. The precision rate is defined as the ratio of correctly detected parts to the sum of correctly detected parts plus false positive. False positive are those regions in the image, which are actually not tampered parts, but have been detected by the algorithm as tampered parts.

$$\text{Precision Rate} = \frac{\text{Correctly detected parts}}{\text{Correctly detected parts + False Positives}} *100 \qquad (6)$$

The Recall rate is defined as the ratio of correctly detected parts to the sum of correctly detected parts plus false negatives. False negatives are those regions in the image, which are actually tampered parts, but have been not detected by the algorithm.

$$\text{Recall Rate} = \frac{\text{Correctly detected parts}}{\text{Correctly detected parts + False Negatives}} * 100\% \qquad (7)$$

Equation (6) and (7) helps to predict the precision and recall rate over various photo edition tools like Photoshop and Microsoft Paint image editor. Table1 explain the data set for computation by considering 100 natural images taken from Sony and Canon digital camera.

Table 1: Computed Value of precision Rate and Recall rate of various Photo Editing Tools.

| Test Data | Number of Images | Precision Rate | Recall Rate |
|---|---|---|---|
| Using Paint | 50 | 92.2 | 92.6 |
| Using Photoshop | 50 | 45.4 | 37.7 |
| Total | 100 | 68.8 | 65.2 |

Table 2. Proposed Detection Methods and Tested Image Format

| Proposed Algorithm | RGB | Gray Scale | JPEG | PNG | BMP |
|---|---|---|---|---|---|
| JPEG Block - Technique | X | X | * | X | X |
| Direction Filter Technique | * | * | * | * | * |

*-→ Successful positive Prediction Of Tampering
X-→ Successful Negative Prediction Of Tampering

Table (1) and Table (2) gives the complete picture of various image formats and computing the prediction of precision rate and recall rate for forgery detection region along with comparison between them in shown in Table (3).

|    | **Attribute Evaluator**     | **JPEG- Block Analysis**                                                                                                | **Direction Filter Technique**                                      |
|----|-----------------------------|-------------------------------------------------------------------------------------------------------------------------|---------------------------------------------------------------------|
| 1. | Image Formation             | Prediction of image tampering with only compression images                                                              | Prediction support Both compression and uncompress ion images       |
| 2. | Time Complexity             | A crafty individual, who wants to perfect an image forgery, with time not a factor, can usually give any detection method trouble | It take less time to predict the forgery region                     |
| 3. | Multiple Forgery Region     | Supports copy-create type forgery                                                                                       | Support Copy-Move Type Forgery                                      |
| 4. | Transformation of Images    | Fails to determine forged anomalies                                                                                     | Successfully determine the forged anomalies                         |

## 5. CONCLUSIONS

This paper focuses on methods to detect digital forgeries created from multiple images called as copy-create image forgeries. Some forgery images that result from portions copied and moved within the same image to "cover-up" something are called as copy-move forgeries. Therefore, the experimental design and analysis herein focuses on copy-create and copy-move image forgeries.

A crafty individual, who wants to perfect an image forgery, with time not a factor, can usually give any detection method trouble. If image tampering occurs in a compressed then JPEG Block methodologies is to support and predict forgery region along with different image format at the same time uncompressed image and then that image is converted to the JPEG image format, the JPEG Block Technique will fail to capture evidence of tampering. This conversion process destroys all proof of tampering since the original tampering does not affect any JPEG blocks. Additionally, any image tampering performed on an image prior to an image size reduction will eliminate detectable anomalies for the direction filter technique

The remainder of the test images returns definitive signs of image tampering when using the JPEG Block Technique for analysis. This method captures the forged area after using various threshold values for testing. The larger threshold value effectively filters out the false positives caused by edges since tampering with an area on the image usually causes greater variability in the JPEG blocks. Consequently, if no pattern arises using different threshold values, the image is most likely authentic or requires analysis by other methods. Overall, the JPEG Block Technique shows promise when used to test an image for tampering. A multifaceted approach is the best practice to follow to decide if an image is forged or authentic when direction filter is used as evidence of tempering.

## REFERENCES

[1] Baxes, G. A., Digital Image Processing: Principles and Applications. New York: John Wiley & Sons, Inc, 1994.
[2] www.fourandsix.com/photo-tampering-history/category/2012
[3] http://ivms.stanford.edu/~varodayan/mentorship/GargHailuSridharan.pdf
[4] A. C. Popescu and H. Farid, "Exposing Digital Forgeries by Detecting Traces of Re-sampling," IEEE Trans. on Signal Processing, vol. 53,no.2, pp. 758–767, Feb. 2005
[5] T. Ng, S.F. Chang, and Q. Sun, "Blind Detection of Photomontage using Higher Order Statistics", IEEE International Symposium on Circuits and Systems, Canada, May 2004
[6] A. C. Popescu and H. Farid, "Exposing Digital Forgeries by Detecting Traces of Re-sampling," IEEE Trans. on Signal Processing, vol. 53,no.2, pp. 758–767, Feb. 2005.
[7] S. Murali, Anami Basavaraj S, and Chittapur Govindraj B. " Detection of Digital Imager forgery Using Luminance Level Techniques", IEEE Third National Conference of Computer Vision, Pattern Recognition, Image processing and Graphics, NCVPRIPG 2011 pp.215.
[8] Fredric, J. and J. Lukas, "Estimation of Primary Quantization Matrix in Double Compressed JPEG Images." Proceedings of DFRWS 2003. Cleveland, OH, August 2003.
[9] E.S.Gopi, N Lakshmanan, T.Gokul and S.Kumar Ganesh "Digital image forgery detection Using artificial Neural Network and Auto Regressive Coefficients" EEE/CCCGET, Ottawa, May 2006 1-4244-0038-4 2006.
[10] Y.Sutch, B.Coskum and H.T.Sencar,N. Memon "Temper Detection Based On Regularities Of Wavelet transform Coefficients " Polytechnic University,Electrical Computer Engineering Dept., Brooklyn Ny 11201, USA
[11] Saha, S. "Image Compression – from DCT to Wavelets: A Review," May 28, 2004 http://www.acm.org/crossroads/xrds6-3/sahaimgcoding.html.
[12] Fridrich, J. and J. Lukas, "Estimation of Primary Quantization Matrix in Double Compressed JPEG Images." Proceedings of DFRWS 2003. Cleveland, OH, August 2003.
[13] Fridrich, J., R. Du, and M. Goljan, "Steganalysis Based on JPEG Compatibility," Special session on Theoretical and Practical Issues in Digital Watermarking and Data Hiding, Multimedia Systems and Applications IV. Pp. 275-280. Denver, CO, August 2001.
[14] Johnson, R. C. "JPEG2000 Wavelet Compression Spec Approved," August 18, 2004. http://www.us.design-reuse.com/news/news1917.html.
[15] A. Garg, A. Hailu, and R. Sridharan "Image Forgery Identification using JPEG Intrinsic Fingerprints" Authors copy
[16] Murali S.,Anami B. S, Chittapur G. B. "Digital Photo Image Forgery Techniques" International Journal Of Machine Intelligence ISSN: 0975-2927 & E-ISSN: 0975-9166, Volume 4, Issue 1, 2012, pp.-405.
[17] James F. O'Brien and hany farid "Exposing Photo Manipulation with Inconsistent Reflections" ACM Transactions on Graphics, Vol. 31, No. 1, Article -, Publication date: January 2012.
[18] Hany Farid and Mary J. Bravo" Perceptual Discrimination of Computer Generatedand Photographic Faces"Digital Investigation 2011
[19] Eric Kee and Hany Farid "A perceptual metric for photo retouching"PNAS Proceeding of the National academy of science of the united state of america doi=10.1073/pnas.1110747108 PNAS November 28, 2011
[20] E. Kee, M. K. Johnson and H. Farid" Digital image Authentication from JPEG headers" IEEE Transactions on Information Forensics and Security, 6(3):1066-1075, 201
[21] E. Kee and H. Farid" Exposing Digital Forgeries from 3-D Lighting Environments"IEEE International Workshop on Information Forensics and Security, 2010
[22] V. Conotter, G. Boato and H. Farid" Detecting Photo Manipulation on Signs and Billboards" International Conference on Image Processing, Hong Kong, 2010
[23] H. Farid and M.J. Bravo" Photo Forensics: How Reliable is the Visual System?" Vision NSciences (VSS), Naples, FL, 2010
[24] E. Kee and H. Farid" Detecting Photographic Composites of Famous People" Technical Report, TR2009- 656, Dartmouth College, Computer Science Hanover, NH 03755
[25] H. Farid "Seeing Is Not Believing" IEEE Spectrum, 46(8):44-48, 2009
[26] W. Wang and H. Farid" Exposing Digital Forgeries in Video by Detecting Double Quantization" ACM Multimedia and Security Workshop, Princeton, NJ, 2009